\documentclass[conference]{IEEEtran}
\IEEEoverridecommandlockouts
\ifCLASSINFOpdf
  \usepackage[pdftex]{graphicx}
\else
   \usepackage[dvips]{graphicx}
\fi
\usepackage{amsmath}
\usepackage{amssymb}
\ifCLASSOPTIONcompsoc
  \usepackage[caption=false,font=normalsize,labelfont=sf,textfont=sf]{subfig}
\else
  \usepackage[caption=false,font=footnotesize]{subfig}
\fi

\usepackage{stfloats}
\usepackage{url}
\usepackage{multirow}
\usepackage{graphicx}

\def\BibTeX{{\rm B\kern-.05em{\sc i\kern-.025em b}\kern-.08em
    T\kern-.1667em\lower.7ex\hbox{E}\kern-.125emX}}
\usepackage[printonlyused,nohyperlinks,nolist]{acronym}
\usepackage{algorithm}
\usepackage{algpseudocode}
\usepackage{cleveref}

\usepackage[numbers,sort&compress]{natbib}

\usepackage{dsfont}

\usepackage{siunitx}
\usepackage{import}
\usepackage{color}

\DeclareUnicodeCharacter{2212}{-}

\begin{document}
%------------------------------------------------------------------
% Acronyms
% use:  \ac{BS}         to use a acronym (first time = full name, then only acronym)
%       \acp{BS}        use the plural
%       \acf{BS}        print the full name and ignore previous declaration
%       \acs{BS}        Use the acronym, even before the first corresponding \ac command
%       \acl{acronym}   Expand the acronym without using the acronym itself.
%       \acresetall     Reset all acronyms (useful after abstract)
%------------------------------------------------------------------

\begin{acronym}
    \acro{lte}[LTE]{Long Term Evolution}
    \acro{mrpe}[MRPE]{mean relative percentage error}
    \acro{ml}[ML]{machine learning}
    \acro{snr}[SNR]{Signal to Noise Ratio}
    \acro{rsrp}[RSRP]{Reference Signal Receive Power}
    \acro{rsrq}[RSRQ]{Reference Signal Received Quality}
    \acro{rssi}[RSSI]{Received Signal Strength Indicator}
    \acro{ue}[UE]{user equipment}
    \acro{qos}[QoS]{quality of service}
    \acro{pqos}[pQoS]{predictive quality of service}
    \acro{phy}[PHY]{physical}
\end{acronym}

\title{QoS prediction in radio vehicular environments via prior user information
\thanks{This work was supported by the Federal Ministry of Education and Research
(BMBF) of the Federal Republic of Germany as part of the AI4Mobile project
(16KIS1170K). The authors alone are responsible for the content of the paper.}}

\author{\IEEEauthorblockN{Noor Ul Ain\IEEEauthorrefmark{1},
Rodrigo Hernang{\'o}mez\IEEEauthorrefmark{1},
Alexandros Palaios\IEEEauthorrefmark{2},
Martin Kasparick\IEEEauthorrefmark{1}
and
S{\l}awomir~Sta{\'n}czak\IEEEauthorrefmark{1}\IEEEauthorrefmark{3}}
\IEEEauthorblockA{\IEEEauthorrefmark{1}Fraunhofer Heinrich Hertz Institute, Berlin, Germany, \{firstname[.middlename].lastname\}@hhi.fraunhofer.de}
\IEEEauthorblockA{\IEEEauthorrefmark{2}Ericsson Research, Herzogenrath, Germany,
alex.palaios@ericsson.com}
\IEEEauthorblockA{\IEEEauthorrefmark{3}Technical University of Berlin, Berlin, Germany}}

\maketitle

\begin{abstract}

Reliable wireless communications play an important role in the automotive industry as it helps to enhance current use cases and enable new ones such as connected autonomous driving, platooning, cooperative maneuvering, teleoperated driving, and smart navigation. These and other use cases often rely on specific \ac{qos} levels for communication.
Recently, the area of \ac{pqos} has received a great deal of attention as a key enabler
to forecast communication quality well enough in advance. However, predicting \ac{qos} in a reliable manner is a notoriously difficult task. In this paper, we evaluate ML tree-ensemble methods to predict QoS in the range of minutes with data collected from a cellular test network. We discuss radio environment characteristics and we showcase how these can be used to improve ML performance and further support the uptake of ML in commercial networks. Specifically, we use the correlations of the measurements coming from the radio environment by including information of prior vehicles to enhance the prediction of the target vehicles. Moreover, we are extending prior art by showing how longer prediction horizons can be supported.
\end{abstract}
\acresetall

\begin{IEEEkeywords}
vehicular networks, radio environments,
quality of service prediction,
machine learning,
test drive

\end{IEEEkeywords}

\section{Introduction}

In the field of vehicular communications, new use cases like connected autonomous driving, platooning, cooperative maneuvering, teleoperated driving, and smart navigation have emerged recently \cite{kulzer_ai4mobile_2021,figueiredo_towards_2001}. Provisioning of these real-time applications requires reliable service quality and uninterrupted connectivity.  However, in wireless applications with high user mobility, \ac{qos} can change drastically within a short period \cite{9129382}. A key enabler for these applications is \ac{pqos}, which has been facilitated by the increasing capabilities of \ac{ml} algorithms \cite{5gaa2019making}. In particular, the problem of data rate estimation has attracted significant attention recently. In literature, instantaneous uplink and downlink data rate estimation has been studied \cite{sliwa_empirical_2019,akselrod_4g_2017,jomrich_enhanced_2019,riihijarvi_machine_2018}, and Schäufele et al. \cite{schaufele_terminal-side_2021} studied the use of quantile estimation methods for predicting bounds on the achievable data rates. Hernangómez et al. \cite{hernangomez_online_2022} studied the non-stationarity of radio environments and proposed an approach based on adaptive online learning for instantaneous \ac{qos} estimation in non-stationary radio environments.

\begin{figure}
    \centering
    \subfloat[Highway radio environment]{\includegraphics[width=0.45\linewidth]{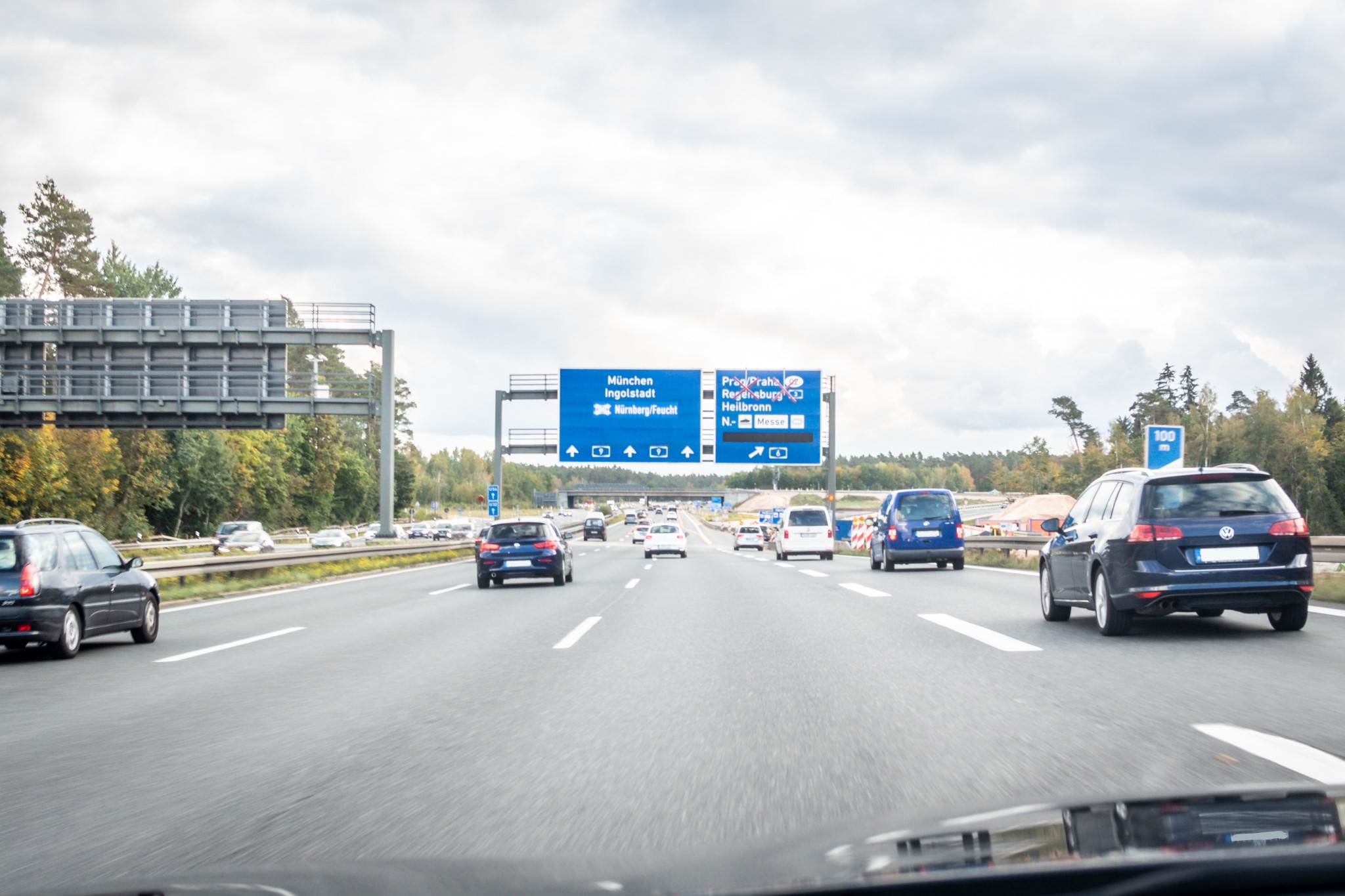}}%
    \hfil
    \subfloat[User Equipment]
    {\includegraphics[width=0.45\linewidth]{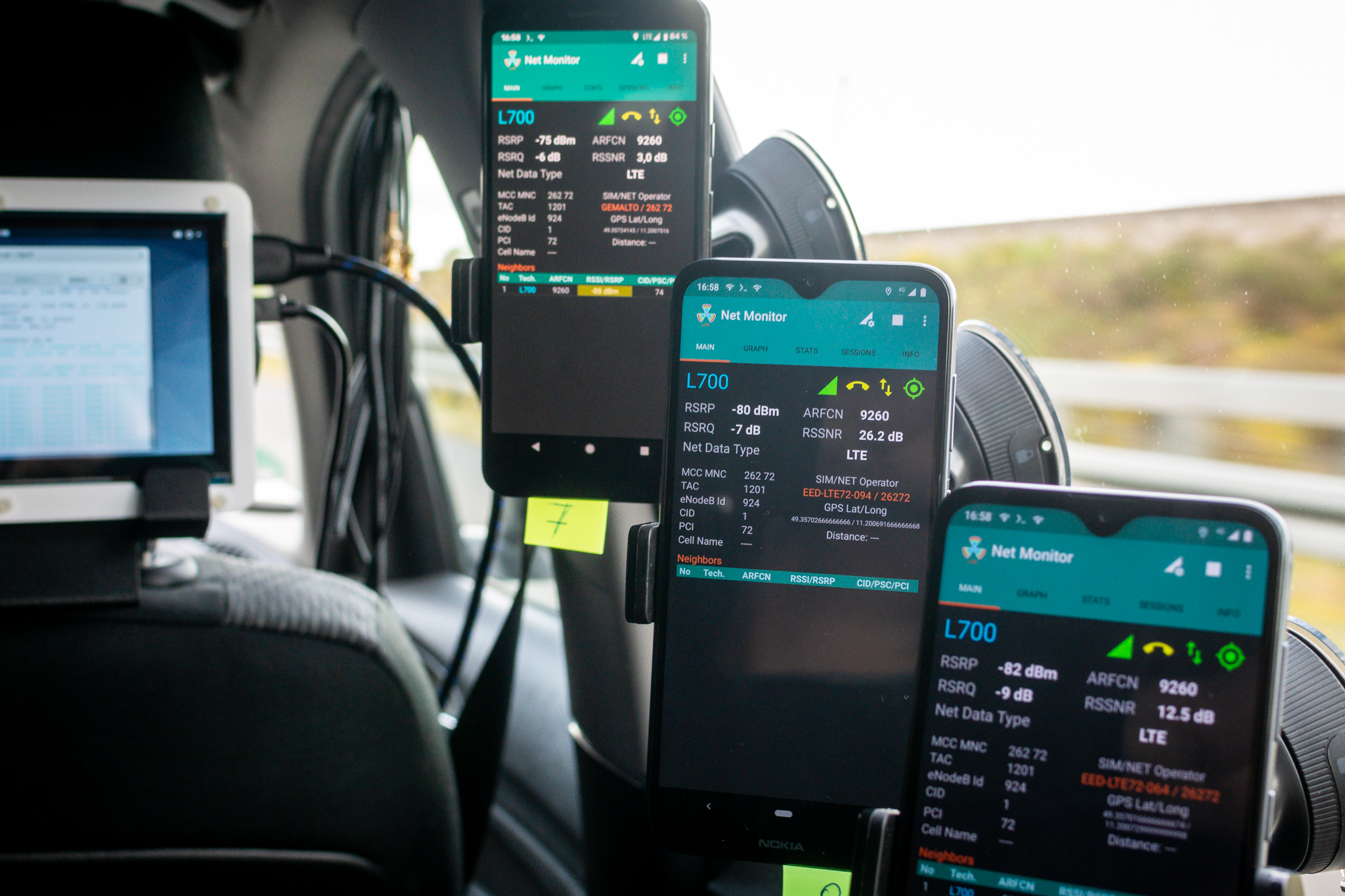}}%
    \hfil
    \subfloat[Measurement monitoring]
    {\includegraphics[width=0.45\linewidth]{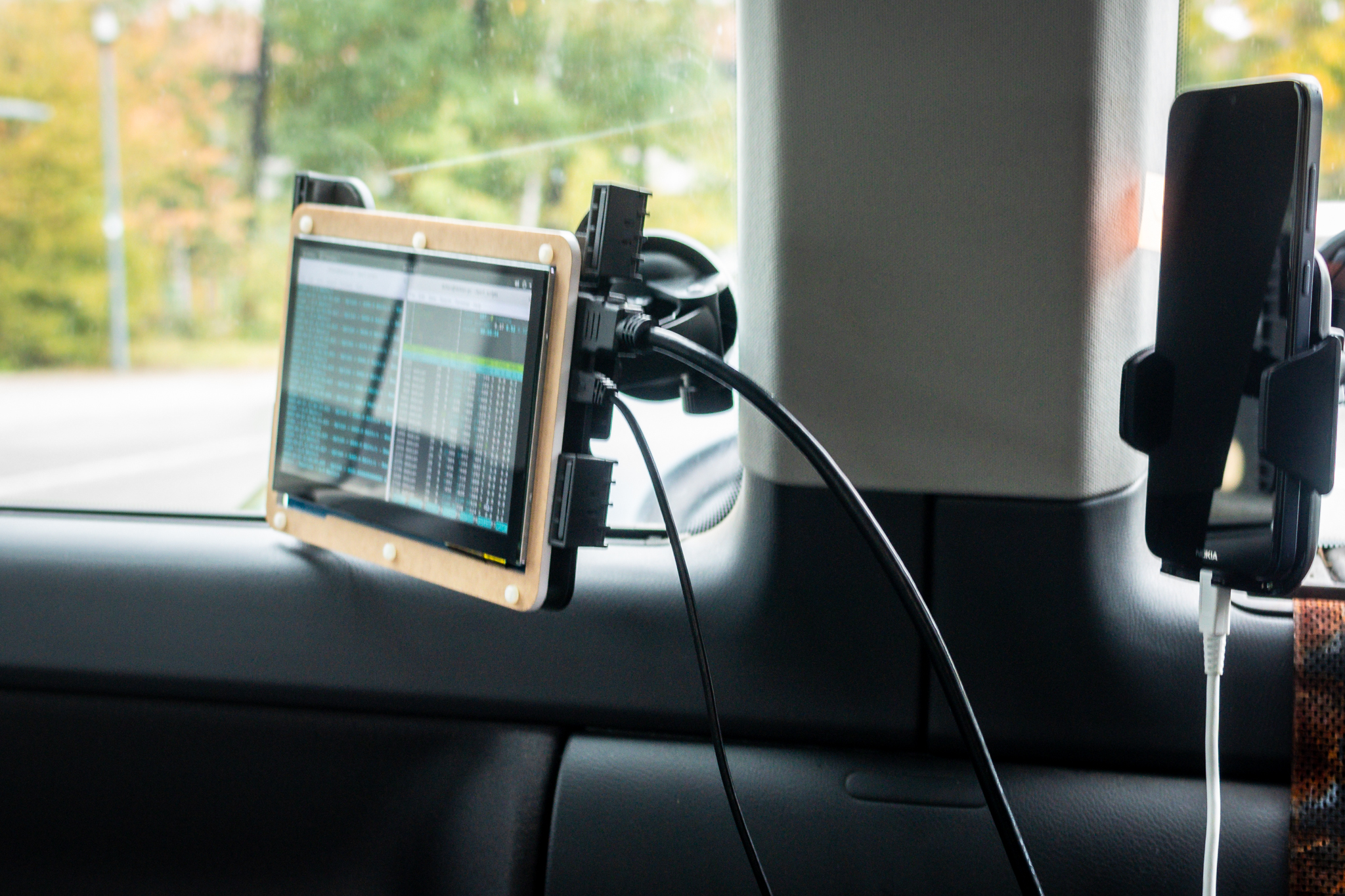}}%
    \hfil
    \subfloat[GPS and LTE antennas]{\includegraphics[width=0.45\linewidth]{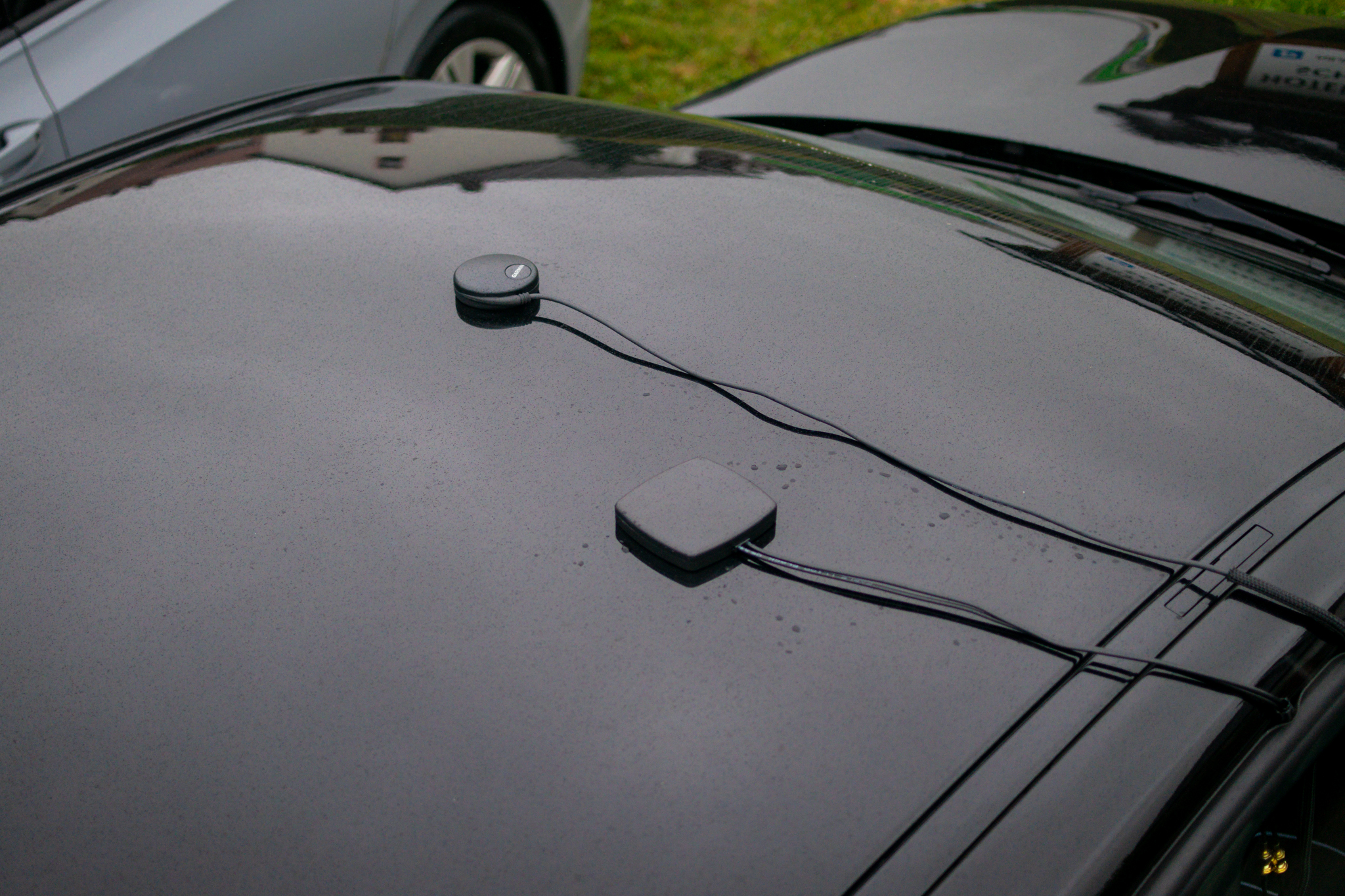}}%
    \caption{``Motorway A9 5G-ConnectedMobility'' measurement
campaign.}
    \label{fig:RadioEnvironmentsFigure}
\end{figure}

In addition to instantaneous \ac{qos}, use cases like teleoperated driving require the estimation of future \ac{qos} in advance. Such a look-ahead \ac{qos} estimation, more commonly referred to as \ac{qos} prediction, can help to avoid any unexpected service interruptions by allowing the network to take actions in a proactive manner. It can also allow safety decisions like the adjustment of velocities or the preparation of safe stop maneuvers in case of insufficient connectivity expected ahead.

In this work, we propose an \ac{ml}-driven approach for \ac{qos} prediction. For instantaneous \ac{qos} estimation, the rich contextual information from the cell and \ac{phy} layer features can significantly improve the results \cite{hernangomez_online_2022,palaios_effect_2021,schaufele_terminal-side_2021}. For predicting future \ac{qos}, however, the information from the cell and \ac{phy} layer is not available. As motivated by \cite{palaios_effect_2021}, we study the correlations for the vehicles following the same route. The results show that \ac{phy} layer measurements exhibit strong correlations with the measurements from the leading vehicles and they undergo similar large-scale fading effects. This raises the question to which extent the information of the past user experience can be useful in predicting the propagation environment. To answer this question, we not only evaluate our ML model for \ac{qos} prediction using the current user information, but we go beyond the state-of-the-art by incorporating information from the past users. We try to overcome the unavailability of \ac{phy} and cell features for \ac{qos} prediction by augmenting our \ac{ml} model with features from leading vehicles as prior information. The use of similar prior information is also studied by Blasco et al. \cite{10032072} for performing \ac{pqos}. We compare the achieved results using different feature sets from the serving cell and the \ac{phy} layer as experienced by the vehicle of interest and the leading vehicles. As a result, we observe that the prior information from leading vehicles results in a better accuracy for \ac{qos} prediction.
Throughout the paper, we refer to our vehicle of interest for which we are predicting \ac{qos} as ``self-vehicle'', and to any leading vehicle as ``next vehicle''. 

The rest of the paper is organized as follows. Section ~\ref{Measurement Campaign} covers the testbed and the observed correlations in \ac{phy} layer features of the vehicles. Section ~\ref{Problem description} introduces the problem description and proposed methodology for \ac{qos} prediction. Section ~\ref{Experiments} covers a comparison of the achieved results using different feature sets and the evaluation of the prediction results at different look-ahead times. Section ~\ref{Conclusion} concludes the findings from our work and proposes potential directions for future work.

\section{Measurement Campaign}
\label{Measurement Campaign}
\begin{figure*}[ht!]
    \centering
    \includegraphics[width=\linewidth]{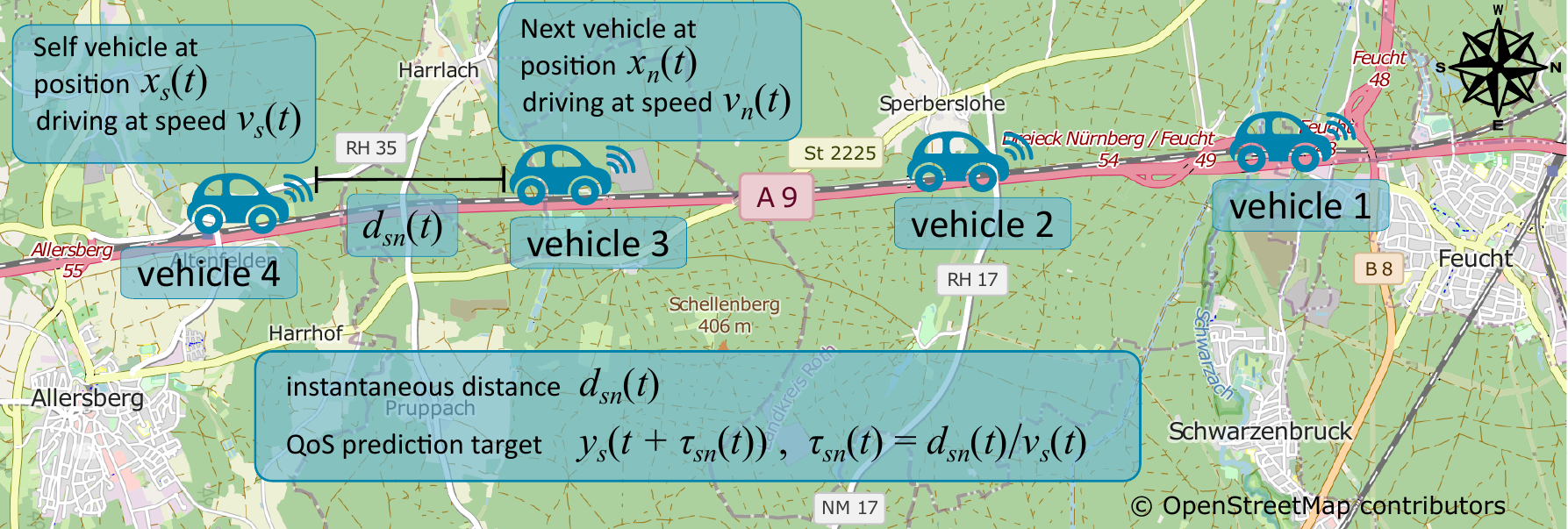}
    \caption{Driving scenario with four vehicles following the same route on highway.}
    \label{fig:scenario}
\end{figure*}
This work uses captured data from the AI4Mobile measurement campaign “Motorway A9 5G-ConnectedMobility” \cite{palaios_network_2021}. This campaign covers data from three radio environments, namely highway, suburban, and rural. The measurements are collected from four \acp{ue} in four different vehicles following the same route at specific time gaps. Each vehicle is equipped with a \ac{ue}, measurement monitoring device and GPS and LTE antennas, as illustrated  in \Cref{fig:RadioEnvironmentsFigure}. The order of the vehicles in our measurements can be seen in \Cref{fig:scenario}. We label these vehicles as vehicle 1, vehicle 2, vehicle 3, and vehicle 4, with vehicle 1 being the leading vehicle and the others following in the same order as listed. The vehicles start their trip at time gaps of nearly 3 minutes each. These time gaps vary to some extent during the round trip as the drivers face slightly different driving conditions. This makes the total coverage span nearly 8 minutes on average, between the first and last vehicle at any point in time. More information about the testbed, the measurement procedure, and the captured parameters can be found in \cite{palaios_network_2021}. 

\begin{figure}[h!]
    \centering
    \includegraphics[width=\linewidth]{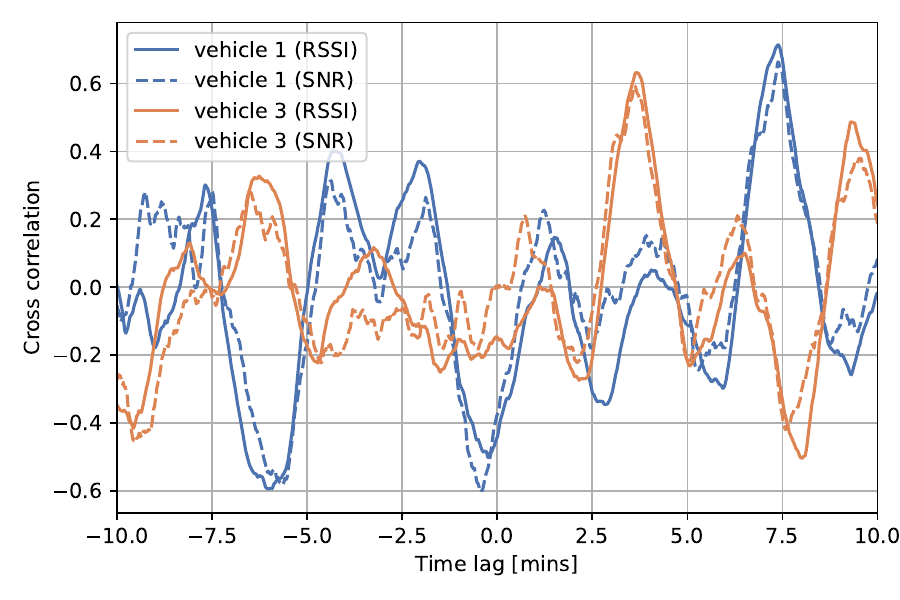}
    \caption{Cross-correlation of self-vehicle with either vehicle 3 or vehicle 1 traveling at $\sim 3$ and  $\sim8$ minutes gap respectively.}
    \label{fig:corr_pc4_pc1}
\end{figure}

For this work, we primarily focus on measurements from the highway, which covers a total distance of approximately 18 kilometers with 6 cells and 3 base stations along the highway. We start by looking at the information of previous users and how this relates to the current user experience. For this, we calculate the correlation in \ac{phy} layer features of two vehicles by taking the time series data of one highway round. \Cref{fig:corr_pc4_pc1} shows the Pearson's cross-correlation of vehicle 4 as self-vehicle with either vehicle 3 (in orange) or vehicle 1 (in blue), over a time interval of [-10,10] minutes. In \Cref{fig:corr_pc4_pc1}, the peak correlation between vehicle 1 and self-vehicle is at nearly 7.5 minutes, corresponding nearly to the average time lag between these two vehicles. The same results can be seen for vehicle 3 having its correlation peak with self-vehicle at almost 3 minutes. This probably indicates that the features from the leading vehicles can be a valuable source of information in \ac{qos} prediction for a vehicle lagging behind it.

\section{Problem description}
\label{Problem description}
For \ac{qos} prediction, we consider the downlink \ac{qos} as our target variable and the features we use convey information both from the cell and \ac{lte} measurements including \ac{snr}, \ac{rsrp}, \ac{rsrq}, and \ac{rssi}. In general, we represent the set of features measured at time $t$ as $x(t)$, which is the input to our \ac{ml} model. For look-ahead time period $\tau>0$, our target variable becomes  $y(t+\tau)\subset\mathds{R}_{+}$ , accounting for \ac{qos} experienced by the vehicle after some time $\tau$. The regression task of \ac{qos} prediction can be modeled as
\begin{equation}\label{eq:1}
 y'(t+\tau)= \hat{f}(x(t),\theta)\;,
\end{equation}
where $y'(t+\tau)$ is the predicted value of \ac{qos} after some look-ahead time period $\tau$, $\theta$ is a set of hyperparameters and $\hat{f}$ is our \ac{ml} model with learned parameters that we obtain as a result of training on the dataset. Note that the information $x(t)$ that we input to our model is the feature set from the self-vehicle at time $t$. However, it is known that the radio environment can change drastically within a few meters \cite{bernado_-_2012}, which calls for the inclusion of knowledge about the radio environment from the expected position of the self-vehicle at time $t+\tau$. As motivated by the correlations discussed above, we incorporate the measurements from other vehicles traveling $\tau$ minutes ahead. 

From the measurement data, we consider a self-vehicle constantly traveling at $v_s(t)$ any time $t$ experiences \ac{qos} $y_s(t)$. Further, we consider that the position of the self-vehicle at $t+\tau$ can be estimated from information including $v_s(t)$, the current position at $t$, the planned route, and the traffic conditions. In a real-world scenario, an operator can fetch data from a preceding vehicle that is sufficiently close to the estimated position at $t+\tau$. However, due to the limited number of cars in our dataset, the value of $\tau$ cannot be selected freely. Thus, we
leverage the simplicity of the cars' trajectories in the measurements to calculate the (time-variable) delay between self and next vehicle, $\tau_{sn}$, as
\begin{equation}\label{eq:2}
\quad\tau_{sn}(t)= d_{sn}(t)/v_s(t)\;,
\end{equation}
which turns predictive \ac{qos} into
\begin{equation}\label{eq:3}
 y_s(t+\tau_{sn}(t))\in\mathbb{R}\;.
\end{equation}
Here, $d_{sn}(t)$ is the instantaneous distance between self and the next vehicle and $y_s(t+\tau_{sn}(t))$ is our target variable.
 
\section{Experiments}
\label{Experiments}
For training and evaluation of our model, we prepare two datasets to conduct two sets of experiments. Firstly, we consider vehicle 4 as self-vehicle and vehicle 1 as next vehicle. We calculate the distance $d_{sn}(t)$ between them for each data sample. The value of $\tau_{sn}(t)$ is calculated using \Cref{eq:2}, yielding an average of $\sim 8$ minutes. The same method is followed for the second dataset, where we keep the same self-vehicle and we consider vehicle 3 as the next vehicle. The value of $\tau_{sn}(t)$ for this pair comes out to be, as expected, $\sim 3$ minutes on average. We use throughput as the \ac{qos} metric, which makes the maximum achieved rate at a specific point in time our target prediction variable. We resample the dataset to a period of 1 second for practical purposes. To analyze the contribution of different knowledge sets toward accuracy, we train in total 10 models for 5 different subsets of network features over 2 different datasets. The complete list of feature sets is given in \Cref{tab:table1}.

\begin{table}[h!]
\begin{center}
\caption{Feature sets of each model}
\label{tab:table1}
\begin{tabular}{|p{0.1\textwidth}|p{0.35\textwidth}|}
 \hline
\textbf{Model} & \textbf{Features x(t)}\\
\hline
{Baseline} & {Current \ac{qos} of self-vehicle} \\
 \hline
 {\ac{phy}} & {\ac{snr}, \ac{rsrp}, \ac{rsrq}, and \ac{rssi} from self-vehicle} \\
 \hline
{\ac{phy} $\And$ Cell} & {\ac{snr}, \ac{rsrp}, \ac{rsrq}, and \ac{rssi} from self-vehicle plus cell load and number of devices connected to the same cell as the self-vehicle} \\
 \hline
{Next \ac{phy}} & {\ac{snr}, \ac{rsrp}, \ac{rsrq}, and \ac{rssi} from next vehicle} \\
 \hline
{Next \ac{phy} $\And$ Cell} & {\ac{snr}, \ac{rsrp}, \ac{rsrq}, and \ac{rssi} from next vehicle plus cell load and number of devices connected to the same cell as the self-vehicle} \\
 \hline
\end{tabular}
\end{center}
\end{table}

For this regression task of QoS prediction, we use a decision tree-based
ensemble ML algorithm called XGBoost regressor \cite{chen_xgboost_2016}. XGBoost has performed better than neural networks and random forests in data rate prediction for the same dataset as studied in \cite{schaufele_terminal-side_2021}. For the evaluation of our regression model, we use \ac{mrpe} as an accuracy metric. In theory, \ac{mrpe} is given by

\begin{equation}
\Ac{mrpe}= \frac{\sum_{(x,y)\in{D}}^{} |{\hat{f}(x(t))-y(t+\tau)|}}{\sum_{(x,y)\in{D}}^{}{|y(t+\tau)|}} *100 \%.
\end{equation}

In practice, we replace actual throughput $y$ in the denominator by $\max(y,1 Mbps)$ to avoid numerical issues with small outliers ($<1 Mbps$) present in the dataset.

 \begin{figure}[h!]
    \centering
    \includegraphics[width=\linewidth]{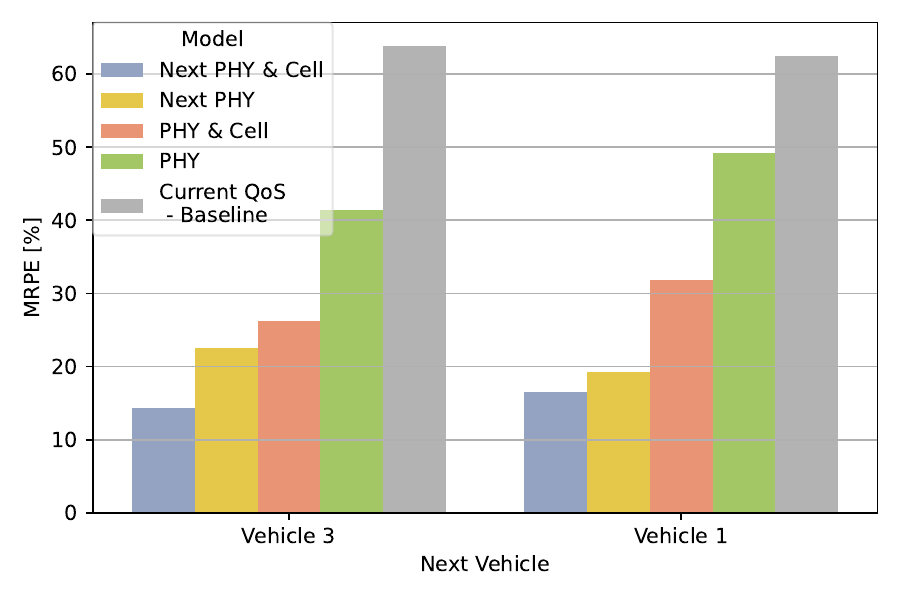}
    \caption{\Ac{mrpe} in \ac{qos} prediction for self-vehicle at look-ahead time 3 and 8 minutes with vehicle 3 and vehicle 1 as next vehicles respectively.}
    \label{fig:percentageError}
\end{figure}

 \Cref{fig:percentageError} shows a bar plot of \ac{mrpe} in \ac{qos} prediction for self-vehicle with vehicle 3 and vehicle 1 as the next vehicle traveling approximately 3 and 8 minutes ahead respectively. The baseline model takes only current \ac{qos} as a feature to predict future \ac{qos}. As we go from right to left sub-bars in each group, the number of input features increases, which results in a drop of \ac{mrpe} at each bar. In \Cref{fig:percentageError}, it can be noticed that the model ``Next PHY $\And$ Cell'' using features from the \ac{phy} layer of the next vehicle and cell values from the self-vehicle's serving cell gives the lowest error, approximately 45\% less than the baseline model.
 
We have performed additional experiments with a bigger feature set including all ``PHY \& Cell'' features from both self and next vehicle, which led to an accuracy drop with respect to ``Next PHY \& Cell''.
Overfitting can be an explanation for this accuracy drop, hence we have run a feature importance analysis~\cite{breiman2001random} across different feature sets to validate this hypothesis. The analysis has revealed that the PHY features of the self-vehicle bear little importance on our prediction task. In contrast, ``Next PHY'' features seem to be the most relevant whenever used, while the importance of the self cell values decreases with the distance to the next car.  
Overfitting, however, depends on the size of the data and thus it cannot be generalized in a real-world scenario. Consequently, we have excluded these additional experiments from the presented results.
 
Moreover, comparing the performance of our prediction models at two different look-ahead times of nearly 3 and 8 minutes, we notice that only the ``Next \ac{phy}'' model has a decrease in \ac{mrpe} as we go from the small to large $\tau$ value for prediction, while all other feature sets linked with self-vehicle have an increase in \ac{mrpe}.

 \begin{figure}[h!]
    \centering
    \includegraphics[width=\linewidth]{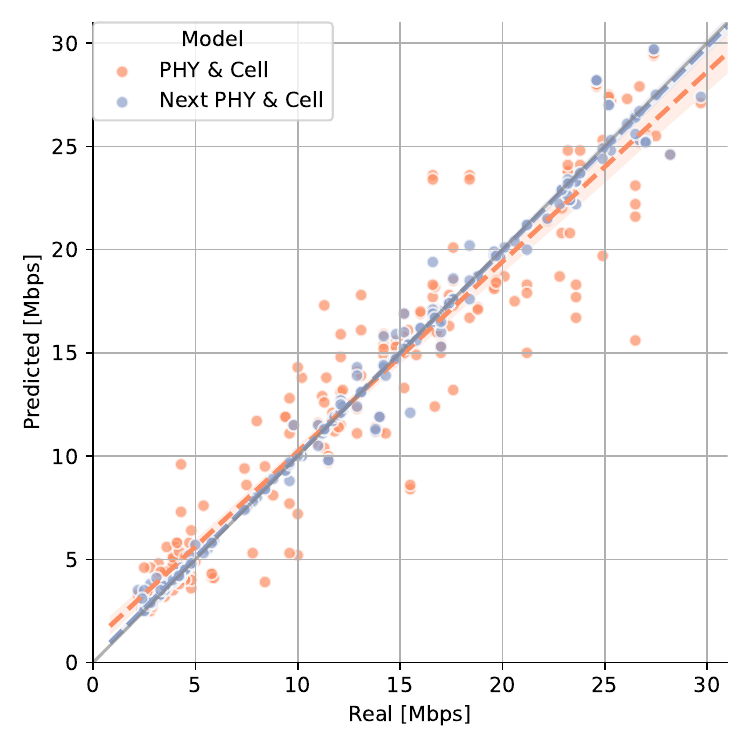}
    \caption{Real($y$) vs. predicted ($\hat{y}$) \ac{qos} comparison between models using \ac{phy} layer features of self-vehicle and next vehicle.}
    \label{fig:scatterplot_3}
\end{figure}

Next, we also perform \ac{qos} prediction for the self-vehicle at the current position of the next vehicle as its future position. Note that $\tau$ takes any value in the range of [0,12] minutes due to crossovers between the vehicles during multiple round trips. However, we observe that the $\tau$ value seems to not have an effect on the performance in a significant way. From \Cref{fig:scatterplot_3}, we see that the next vehicle's \ac{phy} layer features perform significantly better than the model using self-vehicle's features, as the prediction for the ``Next \ac{phy} $\And$ Cell'' model is nearly equal to the diagonal $\hat{y}=y$ (in gray) on the figure.

\section{Conclusion}
\label{Conclusion}

In this work, we have approached the problem of QoS prediction in the context of automotive wireless communication by predicting the maximum achievable throughput in a prediction horizon in the order of minutes. For that, we have leveraged the characteristics  of the radio
environment by incorporating features from distant vehicles,
which improved the estimates about future QoS for the given prediction horizon. This is a first step towards including more information from the radio environment characteristics as captured from preceding vehicles. We were able to show these gains even with only a few vehicles and we believe that there will be even stronger gains once information from more users is used.

We hope that this work serves as a cornerstone to motivate large-scale measurements, and we are positive that these will in turn confirm the benefits of reusing prior information. This can also further optimize the costly data collection campaigns, as less data needs to be collected. In the future, we would like to extend this analysis to larger datasets with many more active users and further look into other aspects of feature engineering for wireless communications, like using the correlation structure of the radio environment. Likewise, we would like to investigate more vehicular environments other than highways, such as rural or urban settings~\cite{hernangomez2023berlin}.

\newpage

\bibliographystyle{IEEEtran}
% argument is your BibTeX string definitions and bibliography database(s)
% {\footnotesize
% \bibliography{references}}
\bibliography{references}

\end{document}